\documentclass[lettersize,journal]{IEEEtran}

\usepackage{stfloats}
\usepackage{amsfonts}
\usepackage{amssymb}
\usepackage{amsthm}
\usepackage{cite}
\usepackage[cmex10]{amsmath}
\usepackage{float}
\usepackage{color}
\usepackage{multirow}
\usepackage[normalem]{ulem}
\usepackage{tabularx}
\usepackage{subfigure}
\usepackage{fancybox,dashbox}
\usepackage{bbm}
\usepackage[dvipsnames]{xcolor}
\usepackage{balance}
\usepackage{algorithm}
\usepackage{algpseudocode}
\usepackage{enumitem}
\usepackage{bm}

\ifCLASSINFOpdf
\usepackage[pdftex]{graphicx}
\DeclareGraphicsExtensions{.pdf,.jpeg,.png}
\else
\usepackage[dvips]{graphicx}
\DeclareGraphicsExtensions{.eps}
\fi

\begin{document}

\title{Goal-Oriented Semantic Communication for Wireless Visual Question Answering}
\author{Sige Liu,~\IEEEmembership{Member,~IEEE,} Nan Li,~\IEEEmembership{Member,~IEEE,} Yansha Deng,~\IEEEmembership{Senior Member,~IEEE,} and Tony Q. S. Quek,~\IEEEmembership{Fellow,~IEEE}
\IEEEcompsocitemizethanks{\IEEEcompsocthanksitem S. Liu, N. Li, and Y. Deng are with the Department of Engineering, King’s College London, London, U.K. (e-mail: sige.liu@kcl.ac.uk; nan.3.li@kcl.ac.uk; yansha.deng@kcl.ac.uk).
\IEEEcompsocthanksitem T. Q. S. Quek is with the Singapore University of Technology and Design, Singapore 487372, and also with the Yonsei Frontier Lab, Yonsei University, South Korea (e-mail: tonyquek@sutd.edu.sg).
}
\vspace{-8pt}}
\maketitle
\begin{abstract}

The rapid progress of artificial intelligence (AI) and computer vision (CV) has facilitated the development of computation-intensive applications like Visual Question Answering (VQA), which integrates visual perception and natural language processing to generate answers. 
To overcome the limitations of traditional VQA constrained by local computation resources, edge computing has been incorporated to provide extra computation capability at the edge side.
Meanwhile, this brings new communication challenges between the local and edge, including limited bandwidth, channel noise, and multipath effects, which degrade VQA performance and user quality of experience (QoE), particularly during the transmission of large high-resolution images. 
To overcome these bottlenecks, we propose a goal-oriented semantic communication (GSC) framework that focuses on effectively extracting and transmitting semantic information most relevant to the VQA goals, improving the answering accuracy and enhancing the effectiveness and efficiency.
The objective is to maximize the answering accuracy, and we propose a bounding box (BBox)-based image semantic extraction and ranking approach to prioritize the semantic information based on the goal of questions. 
We then extend it by incorporating a scene graphs (SG)-based approach to handle questions with complex relationships.
Experimental results demonstrate that our GSC framework improves answering accuracy by up to 49\% under AWGN channels and 59\% under Rayleigh channels while reducing total latency by up to 65\% compared to traditional bit-oriented transmission.
\end{abstract}

\begin{IEEEkeywords}
Goal-oriented, semantic communication, scene graph, visual question answering
\end{IEEEkeywords}
        %



 \vspace{-8pt}
\section{Introduction}

The rapid advancement of artificial intelligence (AI) and computer vision (CV) has driven the development of various computation-intensive applications, increasing the demand for enhanced computational capabilities and performance to meet users' quality of experience (QoE). Visual Question Answering (VQA) is one such application that requires the integration of visual perception and natural language processing to answer a wide range of questions by understanding and reasoning over images and questions\cite{antol2015vqa,lu2016hierarchical}.
Traditionally, VQA tasks have been deployed at local devices such as smartphones, laptops, and unmanned aerial vehicles (UAVs). 
However, these devices encounter significant challenges due to the computational complexity of simultaneously processing large volumes of visual and textual data. This leads to increased computation latency and limited processing power, hindering the efficiency and effectiveness of VQA systems.

To address these challenges, edge computing has emerged as a promising solution by offloading computational tasks from local end devices to edge servers with sufficient computation resource, significantly reducing processing time and latency \cite{shi2016edge,9760192lsg}. 
Meanwhile, this brings new communication challenges between the end devices and the edge, such as limited bandwidth, channel noise, and multipath effects.
As a result, transmitting large volumes of high-quality image data can lead to significant transmission delays, reducing the accuracy of VQA responses and degrading the user quality of experience (QoE).

To tackle these limitations, semantic communication has been introduced in \cite{luo2022semantic}. Unlike traditional image compression techniques, which operate at the pixel level without considering the semantic significance of the data, semantic communication focuses on transmitting only semantically significant information, thereby eliminating redundant data and improving the overall accuracy of VQA tasks \cite{10644029zhouhui}. 

Initial research on semantic communication focused on leveraging deep neural networks (DNN)-based methods to extract the output of the DNN as semantic information from different data modalities (e.g., text\cite{yan2022resource}, audio\cite{weng2021semantic}, images\cite{9928407}, and video\cite{jiang2022wireless}) for efficient transmission. 
On this basis, a DNN-based joint source-channel coding (JSCC) semantic communication architecture has been developed in \cite{bourtsoulatze2019deepjscc}, where the network is jointly trained to reduce the transmission data size by mapping image pixel values to complex-valued channel input symbols.
Subsequently, other advanced JSCC-based methods have been proposed to deal with the limitation of specific channel conditions/environments. 
In \cite{xu2021wireless}, to address the challenge of dynamic channel conditions in wireless communication, the attention mechanism has been leveraged to automatically adapt to various channels during training. This method was further extended to the orthogonal frequency division multiplexing (OFDM) scenario in \cite{9878262}, where the attention mechanism enhanced channel adaptation.
These JSCC-based studies, focused on end-to-end joint training, encounter significant scalability limitations for diverse tasks and fail to accommodate the growing demand for plug-and-play flexibility. 
Their designs primarily focused on channel and coding efficiency by using DNNs to directly compress raw data into the generated feature vectors from the neural network, but they failed to capture and utilize the semantic relevance of the information and lack interpretability of the underlying features.




Meanwhile, recent progress in CV offers new solutions for the semantic representation of images, supporting modular semantic design in VQA.
Based on object detection \cite{diwan2023objectyolo,Girshick_2015_ICCVfasterrcnn,he2017mask}, the bounding box (BBox) of different objects, along with their coordinates and labels, can be leveraged to reason VQA answers \cite{yi2018neural,vatashsky2020vqa}. 
%
To further handle complex relational questions, scene graph (SG) can be extracted from images, representing detected objects as nodes and their relationships as edges \cite{teney2017graph,li2019relation}.
SG generation models, such as those in \cite{zellers2018neuralmotif, yang2018graphrcnn, tang2020unbiased, khandelwal2022iterative}, enable the creation of SGs as a foundation for downstream VQA tasks.
%
These CV methods reduce the size of transmitted data and provide a certain degree of interpretability for semantic information, facilitating their integration into semantic communication. However, several key challenges remain in bridging the gap between semantic communication and CV:
\begin{enumerate}
    \item determining the semantic information to be extracted;
    \item selecting the semantic information to be transmitted;
    \item designing how to extract and transmit semantic information effectively for accurate question answering.
\end{enumerate}




To deal with these challenges, recent studies have started to explore these CV methods in VQA semantic communication. 
In \cite{xie2021task}, the authors employed a convolutional neural network (CNN)-based approach to extract BBox for transmission, replacing the original image. 
\textcolor{black}{In \cite{10255282}, an SG-based semantic communication framework was proposed within a digital twin environment to reduce communication costs and facilitate data transfer for Metaverse services.}
Additionally, a semantic scoring mechanism was introduced in \cite{zhang2023optimizationcmz}, using the concept of image-to-graph semantic similarity (ISS) to rank semantic triplets by considering the frequency and probability of different categories in the dataset. 
However, these approaches are inherently data-centric, focusing exclusively on data-related information during the semantic execution process. Consequently, they struggle to effectively handle diverse questions in VQA, where the same image may be associated with multiple questions that emphasize different aspects and structures.

To address this limitation, we adopt a goal-oriented approach, emphasizing semantic information that is directly related to specific goals instead of focusing solely on datasets or data-related metrics.
Several studies have demonstrated the critical role of goal-oriented methods.
For instance, in robotics \cite{10618994wenchao}, a goal-oriented and semantic communication in robotic control (GSRC) method was proposed to ensure the most important information is utilized. In Metaverse \cite{10577270zhe}, a goal-oriented semantic communication framework in augmented reality (GSAR) method was proposed to extract features from the avatar skeleton graph to prioritize the importance of different semantic information.
However, existing goal-oriented methods are typically designed for specific tasks and scenarios, and are not tailored for VQA in wireless communication contexts. Thus, a dedicated goal-oriented semantic communication framework for edge-enabled VQA is required to ensure flexibility and effectiveness.
%

Inspired by the above, we propose a novel goal-oriented semantic communication (GSC) framework that focuses on effectively extracting and transmitting semantic information most relevant to the VQA goals, thereby improving the answering accuracy and enhancing task effectiveness and communication efficiency.
We develop and evaluate three VQA communication mechanisms, bit-oriented, BBox-based, and SG-based methods, under different channel conditions and QoE requirements. 
To achieve high answering accuracy while maintaining low latency, we provide a comprehensive comparison and practical guidelines for selecting the most suitable VQA communication mechanism. Our contributions can be summarized as follows:
\begin{itemize}
    \item \textcolor{black}{We propose a novel GSC framework tailored for an edge-enabled VQA system over bandwidth-limited, noisy wireless channels. 
    This framework consists of six main modules: a knowledge base for foundational knowledge sharing, a keywords extractor to capture keywords, a question parser to generate the reasoning instructions, a semantic extractor for visual semantic extraction, a semantic ranker to prioritize semantic information, and an answer reasoner to generate accurate responses.}
    \item \textcolor{black}{We formulate the problem as maximizing the average answering accuracy under limited communication bandwidth, characterized by specific fading models, bandwidth constraints, and Signal-to-Noise Ratio (SNR) limitations.} To address this, we leverage a CNN-based BBox generation method to extract the semantic information from image objects and design a goal-oriented BBox (GO-BBox) approach to prioritize the extracted semantic information by ranking BBox based on diverse questions.
    \item 
   To address questions involving complex relationships, we extend our framework by designing an attention-based scene graph (SG) generator to reason on the relationships among objects, followed by a goal-oriented SG (GO-SG) ranking approach that ranks the SGs based on the graph-edit-distance (GED) between images and questions.
    \item We conduct extensive experiments under various channel conditions to compare our proposed GSC framework with the traditional bit-oriented method and state-of-the-art semantic approaches from \cite{liang2021graghvqa} and \cite{zhang2023optimizationcmz}. \textcolor{black}{Our results show that the proposed GSC framework outperforms existing methods}, improving answering accuracy by up to 49\% under AWGN channels and by up to 59\% under Rayleigh channel. Additionally, it reduces overall question execution latency by 65\% compared to bit-oriented transmission.
\end{itemize}

The rest of the paper is organized as follows. Section II presents the system model and problem formulation for VQA. In Section III, we describe the proposed GSC-based BBox semantic ranking method for addressing VQA tasks at the effectiveness level, followed by the extension to the SG semantic ranking method in Section IV. Section V demonstrates the experimental performance evaluation, and finally, Section VI concludes the paper.

\section{System Model and Problem Formulation}
\textcolor{black}{In this section, we introduce our proposed goal-oriented semantic communication (GSC) framework for wireless VQA and then formulate the problem with the goal of achieving high answering accuracy under limited communication resources, characterized by specific fading models, bandwidth constraints, and Signal-to-Noise Ratio (SNR) limitations.}

\subsection{System Model}

\begin{figure*}[t]
    \centering
    \includegraphics[width=0.67\linewidth]{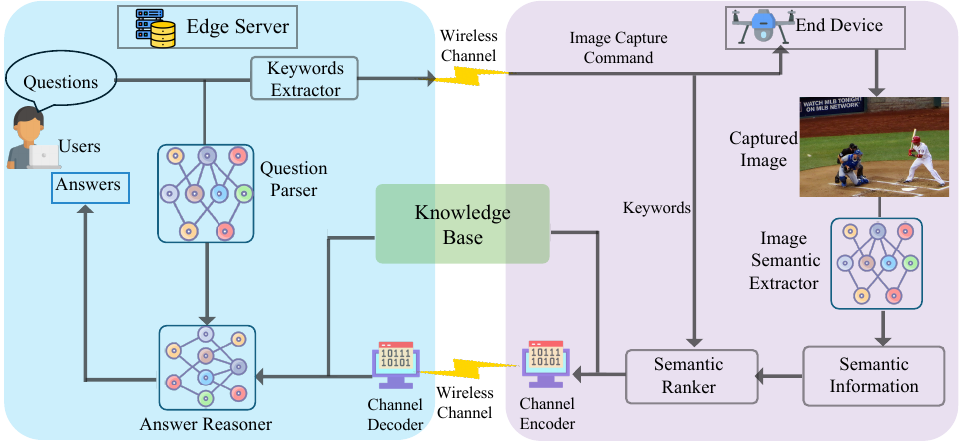}
    \caption{The proposed GSC framework for edge-enabled wireless VQA. }
    \label{fig1}
\end{figure*}

\textcolor{black}{We consider an edge-enabled VQA system operating over a bandwidth-constrained, noisy wireless channel, where human users can query the edge server regarding images captured by end devices, such as unmanned aerial vehicles (UAVs).}
As illustrated in Fig.~\ref{fig1}, at the start of the process, human users initiate VQA tasks by posing questions to the edge server, 
which subsequently instructs the UAV camera to extract semantic information from the captured image and transmit it back via the uplink channel.



\textcolor{black}{We define $Q_{k}$ as the sequence representing the $k$-th question sentence that is raised at the edge server, expressed as
\begin{equation}
    Q_{k}=(q_1,q_2,...,q_{l_k}), k \in [1,N_Q],
\end{equation}
where each $q_i \in Q_{k}$ indicates a natural language word,} and $l_k$ and $N_Q$ denote the sentence length of $k$-th question sentence and the total number of questions, respectively. 

Upon receiving the image capture command from the edge server, the end device captures the corresponding image $I_{j_k}$, where $j_k \in N_I$ is the image index, and $N_I$ represents the total number of images.
It is important to note that VQA typically involves multiple questions associated with a single image. This means the total number of images is generally smaller than the number of questions ($N_I \leq N_Q$), and different questions may refer to the same image ($\exists k \neq k^{'} \text{ s.t. } j_k = j_{k^{'}}$).




The semantic information extractor processes the captured image, which can be expressed as
\begin{align} 
\Psi_{j_k} = f_{ \text{SE}}(I_{j_k}), \label{psi_j_k}
\end{align} 
where $\Psi{j_k}$ represents the extracted semantic information, and $f_{ \text{SE}}(\cdot)$ denotes the semantic extractor for images.
Next, the extracted semantic information $\Psi_{j_k}$ is ranked from a goal-oriented perspective according to the corresponding question $Q_k$. This ranking process is defined as
\begin{align} 
\Psi^R_{j_k} = f_{\text{SR}}(\Psi_{j_k} ; Q_k), 
\end{align} 
where $\Psi^R_{j_k}$ denotes the ranked semantic information, and $f_{\text{SR}}(~\!\!\cdot ~\!\!; Q_k)$ represents the semantic ranking operation based on the question $Q_k$.

Subsequently, the ranked semantic information can be tokenized using the knowledge base to save communication resource, after which the tokenized information is encoded and transmitted back to the edge server via the uplink wireless channel. 
Last, the answer reasoner at the server receives both the decoded image semantic information and the instructions steps from the question parser module and generates an answer for the user.

\subsection{Communication Model}

In order to focus on the uplink communication of image semantic information from the end device to the edge server, we assume that the downlink transmission of image capture command and question keywords are ideal and free of errors. The uplink communication process starts with source encoding, where the original data is converted into a bitstream, denoted by \( \mathbf{x} \). 
This bitstream is then encoded using Low-Density Parity-Check (LDPC) coding \cite{chen2005reduced} for error correction. LDPC codes are characterized by a sparse parity-check matrix \( H \). The LDPC encoding ensures that the generated codeword \( \mathbf{c} \) satisfies the parity-check condition \( H \mathbf{c}^T = \mathbf{0} \).
After that, the encoded bitstream \( \mathbf{c} \) is subjected to binary phase-shift keying (BPSK) modulation. This modulation technique alters the phase of a carrier signal according to the encoded bits, producing a modulated signal represented as \( \mathbf{y} \), which is then transmitted over the wireless channel. 
For each binary input $c_i\in \mathbf{c}$, the corresponding modulated output $y_i\in\mathbf{y}$ can be expressed as
\begin{align}
    y_i = \left\{
\begin{array}{ll}
+1& \text{if } c_i  = 1, \\
-1 & \text{if } c_i = 0.
\end{array}
\right.
\end{align}
\begin{figure}
    \centering
    \includegraphics[width=0.75\linewidth]{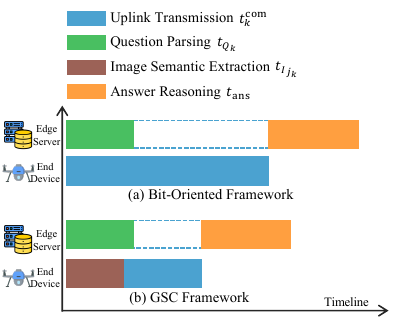}
    \caption{The latency comparison between bit-oriented framework and GSC framework.}
    \label{fig_add1}
\end{figure}

\begin{figure*}
    \centering
    \includegraphics[width=0.91\linewidth]{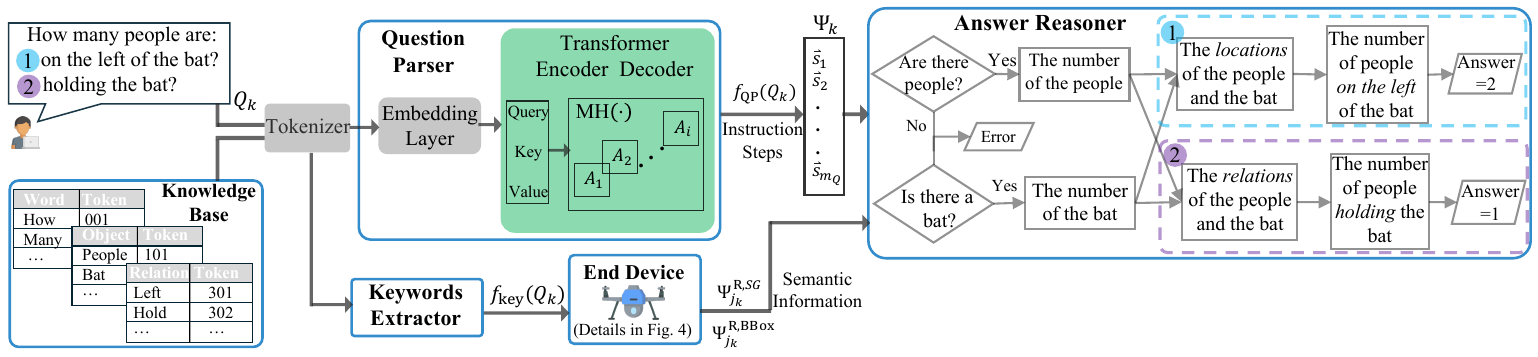}
    \caption{\textcolor{black}{Workflow at the edge server in the GSC framework, including keywords extractor, question parser, and answer reasoner.}}
    \label{fig_question}
\end{figure*}

The received binary signal $\mathbf{y}'$ can be expressed as 
\begin{align}
    \mathbf{y}' = \mathbf{y}\otimes h_k + \mathbf{n}_k,
\end{align}
where $\mathbf{n}_k$ represents the channel noise.  The signal is then demodulated, followed by LDPC decoding, to recover the original data.

The Signal-to-Noise Ratio (SNR) at $k$-th quesiton $Q_k$ can be expressed as
\begin{equation}
    \text{SNR}_k=\frac{P_k \Vert h_k \Vert^2 }{N_0},\label{snr}
\end{equation}
where $P_k$ is the transmission power of the end device, $h_k$ is the small-scale fading coefficient, and $N_0$ is the noise power. 

On this basis, the communication latency for bitstream transmission from the end device transmitting selected semantic features to the edge server can be expressed as
\begin{equation}
    t_k^{\text{com}} = \frac{c( \Psi^R_{j_k})}{W\log_2(1+\text{SNR}_k)},\label{t_tra}
\end{equation}
where $c( \Psi^R_{j_k})$ is the data size (in bits with float 32 data type) of the transmitted semantic information from the end device to the edge server, and $W$ is the channel bandwidth.
Here, we define the image $I_{j_k} \in \mathbb{R}^{\text{H}\times \text{W} \times \text{D}}$, where H, W, and D represent the height, width, and colour channels of the image, respectively. Each pixel's colour channel is stored as a 32-bit floating point number, and the data size of the image can be calculated by multiplying its spatial dimensions and the number of colour channels, yielding $32\times \text{HWD}$ bits.

The total latency, denoted as $t_k^{\text{total}}$, is defined as the elapsed time from the initiation of a task by the human user to upon receiving the output from the answer reasoner. 
This process requires two distinct inputs: the reasoning instruction steps $\Psi_{k}$ originating from the question parser at the edge, and the semantic information of the image relayed from the end device through a wireless communication channel.
Consequently, the total latency can be expressed as
\begin{equation}
    t_k^{\text{total}}=\max\{ t_{Q_k}, t_{I_{j_k}}\!\!+\!t_k^{\text{com}} \} + t_{\text{ans}}, \label{total_latency}
\end{equation}
where $t_{Q_k}$,  $t_{I_{j_k}}$, and $t_{\text{ans}}$ denote the processing time of question parser at the edge server, the corresponding image at the end device, and the answer reasoning, respectively. 
The roles of these latencies are illustrated in Fig.~\ref{fig_add1}, where we also compare the traditional bit-oriented framework with our proposed GSC framework in terms of the total latency experienced at the edge server and end device. 
A detailed comparison of latency and answer accuracy is presented in Fig.~9 of the experimental results.



\subsection{Problem Formulation}

The goal of the proposed VQA system is to maximize answering accuracy. Given the \( k \)-th input question \( Q_k \) and its corresponding image \( I_{j_k} \), we denote the answer reasoning process as \( f_{\text{ans}}(Q_k, I_{j_k}) \). Based on this, the answering accuracy maximization problem can be formulated as
\begin{subequations} \label{main_p1}
\begin{align}
    & \mathcal{P}_1: \max \frac{1}{N_Q} \sum_{k=1}^{N_Q} \mathbbm{1} \{ f_{\text{ans}}(Q_k, I_{j_k}) = A_k \}, \label{objective function}\\
    & \text{s.t.}  ~~k \in [1, N_Q],~~ j_k \in [1, N_I], \label{p1_st_1}\\
    & ~~~~~~ W\leq W_0, \label{p1_st_2}
\end{align}
\end{subequations}
where \( A_k \) represents the correct answer to the \( k \)-th question, and the indicator function \( \mathbbm{1}\{ \cdot \} \) returns 1 if the generated answer \( f_{\text{ans}}(Q_k, I_{j_k}) \) matches \( A_k \), and 0 otherwise. 
The constraint in \eqref{p1_st_1} specifies the indices of the question and image involved and the limitation of communication bandwidth \( W_0 \).

It is worth noting that in traditional bit-oriented communication frameworks, communication delay \( t_k^{\text{com}} \) is often significant due to communication resources required to transmit high-resolution images from multiple cameras simultaneously to the edge server.
Consequently, the answer reasoning process can only commence after the complete reception of both the questions and images, which impedes real-time data processing and reduces the overall framework's effectiveness, ultimately affecting the accuracy of the answers.




\section{Natural Language Processing for Questions at the Edge Server}
\textcolor{black}{In this section, we present the detailed framework design and workflow for the natural language processing of questions at the edge server.
At the edge server, the primary task is to process natural language questions, and the workflow for this process is illustrated in Fig.~\ref{fig_question}, which outlines the key modules: the knowledge base, keywords extractor, question parser, and answer reasoner.}
\subsection{\textcolor{black}{Knowledge Base}}
The knowledge base is designed to standardize all question inputs, facilitating consistent interpretation, training, and communication across the framework. 
Operating transparently for both the edge server and the end device, it works in conjunction with the tokenizer to reduce the size of information exchanged between modules, thereby enhancing efficiency and minimizing resource consumption.

To achieve this, we leverage GloVe \cite{pennington2014glove} with 600 dimensions to present all natural language words. Each word is mapped to the closest concept in the predefined vocabulary, minimizing the effects of synonyms, plurality, and tense variations on the interpretation of questions.
As depicted in Fig.~\ref{fig_question}, the knowledge base consists of three tables:  1) all natural words and their tokens, 2) objects and their associated tokens, and 3) relationships and their tokens.
Notably, the knowledge base can dynamically assign new tokens and update the tables when encountering new words, ensuring it remains up to date \cite{10577270zhe}.
\subsection{\textcolor{black}{Keywords Extractor}}
The keywords extractor $f_{\text{key}}(\cdot)$, built upon the knowledge base, is designed to capture the relevant keywords $f_{\text{key}}(Q_{k})$, including the associated image index, objects, and relationships.
The image index is used by the end device to capture and retrieve the correct images, while the objects and relationships will be utilized in Sections IV and V for further goal-oriented semantic ranking.
For the given example question $Q_k$ in Fig.~3, the extracted keywords include the associated image index $j_k$ and the tokenized objects and relationships, as represented in the tables within the knowledge base.
\subsection{Question Parser}\label{section_question}
The question parser is designed to translate questions into a sequence of reasoning instructions, which are passed to the answer reasoner to guide the reasoning process.
We denote the sequence of instruction steps as $\Psi_k$, which can be expressed as
\begin{align} 
\Psi_k = f_{\text{QP}}(Q_k ), \label{psi_k}
\end{align}
where $f_{\text{QP}}(\cdot )$ denotes the question parsing process.

Specifically, we design a multi-layer transformer-based \cite{vaswani2017attention} question parser employing an attention-based sequence-to-sequence model with an encoder-decoder structure. 
The transformer has an encoder-decoder structure and consists of stacked attention functions and the output of each attention head is computed as
\begin{equation}
\text{A}_i(\text{Query, \!Key, \!Value}) \!=\! \text{softmax}\!\left(\!\frac{\text{Query}_i\text{Key}_i^T}{\sqrt{d_k}}\!\right)\!\text{Value}_i,\label{attention}
\end{equation}
where  \( \text{Query }\), \(\text{Key}_i \), \( \text{Value}_i \), and $d_k$ refer to query, key, value vectors and the scaling factor. 
Then, the multi-head attention can be written as
\begin{align}
    \text{MH}(\text{Query, Key, Value}) = [\text{A}_1,...,\text{A}_h]U_O,\label{multihead}
\end{align}
where MH($\cdot$) concatenates the outputs from the single attention heads followed by the projection with trainable parameters $U_O$.
In this paper, we use $l=3$ identical transformer encoder and decoder layers. 
%
The input question $Q_{k}$ is fed into the embedding layer to encode the question sequence into one-hot vectors for easier execution in the subsequent question parser.  
Hence, according to equation \eqref{psi_k} the task parsing process can be written as
\begin{equation}
  \Psi_k = [\vec{s}_1,\vec{s}_2,...,\vec{s}_{m_Q}]  = f_{\text{QP}}(Q_k), \label{after lstm}
\end{equation}
where $m_Q$ denotes the total number of vectors in $\Psi_k$, and $\vec{s}_i$ denotes the true probability of the $i$-th class in the ground truth.
The cross-entropy is used in the loss function $\mathcal{L}_Q$, which can be written as 
\begin{equation}
     \mathcal{L}_Q = \arg \min_{\theta_{\text{QP}}} -\sum_{i=1}^{m_Q} \vec{s}_i \log(\hat{\vec{s}}_i),\label{loss_q}
\end{equation}
where $\theta_{\text{QP}}$ encompasses the network parameters of the question parser, and $\hat{\vec{s}}_i$ is the predicted probability of the $i$-th class. 

\subsection{Answer Reasoner}
The answer reasoner is designed to perform the logical operations to process the questions and generate accurate answers.
It takes two inputs: the reasoning instruction steps $\Psi_k$ and image semantic information $\Psi^R_{j_k}$.

The answer reasoner operates by following a structured logical flow, as illustrated in Fig.~3, to analyze the input information and systematically derive the appropriate response. 
The input reasoning instructions $\Psi_k$ specify the key semantic elements to focus on each question, such as specific objects, relationships, or spatial attributes, and each step $\vec{s}_i \in \Psi_k$ represents an individual operation within the reasoning process.
At each step $\vec{s}_i$, the answer reasoner uses the extracted semantic information to match the relevant semantic elements to the specified instructions. 
After processing a step $\vec{s}_i$, the reasoning result is used as input for the next step $\vec{s}_{i+1}$. This sequential process continues until the final step $\vec{s}_{m_q}$, where $m_q$ is the total number of steps of the given question.
Once all steps are completed, the answer reasoner synthesizes the results to generate an answer to the question.




For the example questions in Fig.~3, the answer reasoner processes the input information as follows: For the first question (``\textit{How many people are on the left of the bar}”), it identifies the relevant locations of the people and the bar, with the positional reasoning process detailed in Section IV. For the second question (``\textit{How many people are holding the bar}”), it focuses on the relationships between the people and the bar, with the relational reasoning discussed in Section V.

\begin{figure*}
    \centering
    \includegraphics[width=0.7\linewidth]{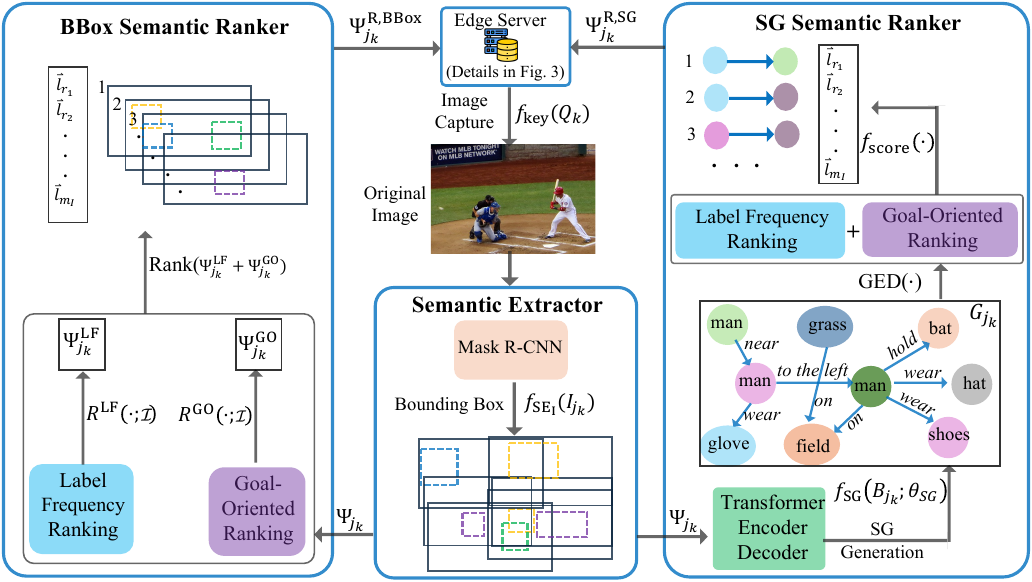}
    \caption{Workflow at the end device in the GSC framework, including the semantic extractor along with the BBox and SG semantic rankers.}
    \label{fig_image}
\end{figure*}

\section{Bounding Box-Based Semantic Processing for Images}
In this section, we present the detailed GSC framework design at the end device and the workflow for the visual processing of images, and provide a goal-oriented bounding box (GO-BBox) algorithm to solve the formulated problem.

\subsection{Image Semantic Extractor}
As illustrated in Fig.~4, we employ a Mask R-CNN \cite{he2017mask} network to segment image regions and predict their categorical attributes such as color, material, size, and shape. 
The extracted image semantic information is aligned with the reasoning requirements defined in equation \eqref{psi_j_k}, enabling goal-oriented ranking and semantic communication.
To this end, the process of the image semantic extractor can be written as
\begin{equation}
   \Psi_{j_k} = [\vec{l}_1,\vec{l}_2,...,\vec{l}_{m_I}]  = f_{\text{SE}}(I_{j_k}), \label{after_cnn}
\end{equation}
where $m_I$ denotes the total number of vectors in $\Psi_{j_k}$, and $\vec{l}_j$ denotes the true probability of the $j$-th class in the ground truth. The cross-entropy is leveraged in the loss function $\mathcal{L}_I$, which can be expressed as 
\begin{equation}
     \mathcal{L}_I = \arg \min_{\theta_{\text{SE}}} -\sum_{j=1}^{m_I} \vec{l}_j \log(\hat{\vec{l}}_j),\label{loss_i}
\end{equation}
where and $\theta_{\text{SE}}$ encompasses the network parameters of the image semantic executor for the object detection, and $\hat{\vec{l}}_j$ is the predicted probability of the $j$-th class. 
%
Therefore, considering the semantic information extracted from questions and images, problem $\mathcal{P}_1$ can be converted as
\begin{subequations} \label{main_p2}
\begin{align}
    & \mathcal{P}_2: \max  \frac{1}{N_Q} \sum_{k=1}^{N_Q} \mathbbm{1} \{ f_{\text{ans}}(\Psi_k, \Psi_{j_k} ;\theta_{\text{SE}},  \theta_{\text{QP}})  = A_k \} , \label{objective function_p2}\\
    & \text{s.t.}  ~~k \in [1, N_Q],~~ j_k \in [1, N_I], \label{p2_st_1}\\
    & ~~~~~~ W\leq W_0.\label{p2_st_2}
\end{align}
\end{subequations}

The communication resources from the end device to the edge server are constrained by limited bandwidth and Varying SNR. Additionally, the stringent latency requirements of VQA further restrict the amount of semantic information that can be transmitted. 
Therefore, it is crucial to rank the semantic information to identify and prioritize the most relevant content for transmission.

\subsection{BBox Semantic Ranker and GO-BBox Algorithm}
\textcolor{black}{To address these challenges, we design the BBox semantic ranker and propose a GO-BBox algorithm to rank the extracted semantic information $\Psi_{j_k}$.
We first utilize a fundamental ranking method in \cite{zhang2023optimizationcmz}, which is designed to organize semantic information extracted from the original image in descending order, based on the different label frequencies (LF) of the objects in the images. After extracting the image semantic information, each vector in $\Psi_{j_k}$ is assigned a score based on its priority relative to image-related data. We define $\Psi_{j_k}^{\text{LF}}$ as the ranked semantic information based on the categories frequency, which can be expressed as}
\begin{equation}
    \Psi_{j_k}^{\text{LF}}= R^{\text{LF}}(\Psi_{j_k}; \mathcal{I}),
\end{equation}
where $\mathcal{I}= (I_{j_k})_{j_k=1,...,N_I}$ indicates the image dataset, and $R^{\text{LF}}(\cdot;\mathcal{I})$ is the ranking process based on the image data set $\mathcal{I}$.
For each vector $\vec{l}_{i}\in \Psi_{j_k}$,  the ranking process can be written as
\begin{align}
    R^{\text{LF}}(\vec{l}_{i}; \mathcal{I}) = \frac{n(\vec{l}_i)}{\sum_{\vec{l}_j\in \mathcal{I}} n(\vec{l}_j)} \label{rank1_2},
\end{align}
where $n(l_i)$ is the category frequency of the semantic feature $l_i$ in the considered image dataset. 
All semantic information is initially scored using the LF ranking algorithm; however, this approach does not consider the correlation between the question and the semantic information extracted from the image.

To address this limitation, we leverage the goal-oriented (GO) concept by utilizing the extracted keywords $f_{\text{key}}(Q_k)$ at the end device. We define $\Psi_{j_k}^{\text{GO}}$ as the GO-ranked semantic information, which can be expressed as
\begin{equation}
    \Psi_{j_k}^{\text{GO}}= R^{\text{GO}}(\Psi_{j_k}^{\text{LF}}; f_{\text{key}}(Q_k)),
\end{equation}
where $R^{\text{GO}}(\cdot;f_{\text{key}}(Q_k))$ is the DO semantic ranking process based on the question-related keywords $f_{\text{key}}(Q_k)$. 
For each vector $\vec{l}_{j}\in \Psi_{j_k}$, we define the online ranking process as
\begin{align}
    R^{\text{GO}}(\vec{l}_{j}, f_{\text{key}}(Q_{k})) = \sum_{j=1}^{N_I} w_j\cdot\mathbbm {1}\{\vec{l}_{j} \in \Psi_{k})\}, \label{rank1_1}
\end{align}
where $w_j\geq 1$ is the weight parameter, and $\vec{V}_{Q}$ is the goal-oriented semantic information obtained from \eqref{after lstm}.
Now we can combine the two score processes for all the image semantic features in a descending order.
As $\Psi_{j_k}^{\text{LF}}\in (0,1), \forall \vec{l}_i$ and $\Psi_{j_k}^{\text{LF}}\geq 1, \forall \vec{l}_{j}$, we can denote the ranked semantic vectors in BBox as 
\begin{equation}
\Psi_{j_k}^{\text{R,BBox}}=\text{Rank} (\Psi_{j_k}^{\text{LF}}+\Psi_{j_k}^{\text{GO}}),\label{ranked}
\end{equation}
where Rank$(\cdot)$ represents a descending sorting operation based on the scores of the BBox.
Then, the top $N_{\text{top}}$ entries in $\Psi_{j_k}^{\text{R,BBox}}$ are transmitted to the edge for answer reasoning, and the workflow of the proposed GO-BBox algorithm is summarized in \textbf{Algorithm 1}.

\begin{algorithm}[t]
\caption{{GO-BBox Algorithm}}
\label{algoritm_rank1}
 \begin{algorithmic}[1]
	\State Input:  $(Q_{k})_{k=1,...,N_Q}$, $(I_j)_{j=1,...,N_I}$.
    \State Input: Pre-trained question parser $\theta_{\text{QP}}$ and semantic extractor $\theta_{\text{SE}}$.
	
    \For{$k = 1,...,N_Q$}
    	\State  Get question sequence $Q_{k}$.
            \State Extract keywords $f_{\text{key}}(Q_{k})$ from keywords extractor.
            \State  Generate the reasoning instruction steps $\Psi_k$ from question parser according to \eqref{after lstm}.
            
            \State  The end device receives the keywords $f_{\text{key}}(Q_{k})$ and captures the corresponding image  $ I_{j_k}$.
            \State  Extract the image semantic information $\Psi_{j_k}$ from the semantic extractor according to \eqref{after_cnn}.

        \If{ $\vec{l}_{i}\in \Psi_{j_k}$}	        

                    \State Get the ranking scores $ R^{\text{LF}}(\vec{l}_{i}; \mathcal{I})$ according to \eqref{rank1_2} based on different LF.

	                \State Get the goal-oriented ranking scores $R^{\text{GO}}(\vec{l}_{i}; \mathcal{I})$ according to \eqref{rank1_1}. 
	       
    \EndIf
            \State Get the ranked semantic BBox $\Psi_{j_k}^{\text{R,BBox}}$ according to \eqref{ranked}.
        
\EndFor

\end{algorithmic}
\end{algorithm}

\section{Scene Graph-Based Semantic Processing for Images}

In this section, we extend our analysis to the VQA task with questions related to complex relationships between objects. 
Considering the two example questions in Fig.~\ref{fig_question}, the previous section primarily addresses the BBox-based semantic ranker, which performs well for questions related to positions or attributes (e.g., the first question) but encounters challenges when dealing with more complex, relationship-based questions (e.g., the second question).
To overcome this limitation, we utilize scene graphs (SG), as shown in Fig.~4, to represent semantic information, where nodes represent objects (e.g., \textit{man}, \textit{horse}, \textit{bus}) and edges denote their relationships (e.g., \textit{ride}, \textit{hold}, \textit{wear}). 
Notably, a BBox can be viewed as a special case of an SG, where relationships are restricted to spatial locations and attributes. Similarly, the analysis in Section IV can be considered a specific instance of the broader framework discussed in Section V


 
\subsection{Goal-Oriented Scene Graph Generation}


We define SG as a representation of real-world entities and the relationships between them. 
Specifically, we define triplet $\gamma_n=<g_s,g_o,g_r>$ to represent the format describing a relationship between two entities in an SG. Here, $g_s$ is the subject of the triplet, $g_o$ indicates the object of the triplet, $g_r$ represents the relationship between the subject and the object. 
For example, if a man is riding a horse in the image, a corresponding triplet in the SG to be generated can be represented as 
\begin{align}
    <g_s=\text{Man}, g_o=\text{Bat}, g_p=\text{Hold}>
\end{align}
Therefore, an SG of an image can be viewed as a combination of $N_{j_k}$ triplets, represented as $G_{j_k}=\{\gamma_1,...,\gamma_{N_{j_k}}\}$ 

To generate an SG, we utilize the Region Proposal Network (RPN) layer of the Mask R-CNN network, as described in Section III, where we define the set of BBox for image $j_k$ as $B_{j_k}$.
The goal of SG generation is to learn the mapping 
\begin{align}
    f: G_{j_k} = f_{\text{SG}}(B_{j_k}; \theta_{\text{SG}})=\{\gamma_1,...,\gamma_{N_{j_k}}\}, \label{after_cnn_sg}
\end{align}
where $f_{\text{SG}}(\cdot; \theta_{\text{SG}})$ represents the SG generation  process, and $\theta_{\text{SG}}$ encompasses the network parameters for the SG generation.  We define $G_{k}$ and $G_{j_k}$ as the SG of the question $k$ and image $j_k$, respectively. 

According to the multi-head attention mechanism in equation \eqref{attention} and \eqref{multihead} in Section \ref{section_question}, we utilize the transformer-based structure to process the BBoxes $B_{j_k}$ to generate the SG. The BBoxes are passed through the transformer encoder to produce contextualized embedding that contains both object-specific information and global scene context. The decoder then operates in an autoregressive manner to generate the next relationship triplet at each time step based on the previously generated triplets.
The Transformer encoder follows a similar architecture to that described in Section \ref{section_question}, with the input modified to use BBox data instead of text.
During the decoding process, the decoder leverages previously predicted relationship triplets at each time step to build a history embedding, representing the current state of predictions. Subsequently, all possible pairs of contextualized object embeddings, excluding those already predicted, are combined with the historical embedding to predict new relationships.

Given the previous $i$ triplets $\{\hat{G}_{j_k}\}_{1:i}$, our goal is to learn the conditional probability of the next triplet, which can be written as 
\begin{align}
    \hat{\gamma}_{i+1} = \arg \max_{\hat{\gamma}} p(\hat{\gamma} \mid B_{j_k}, \{\hat{G}_{j_k}\}_{1:i}).
\end{align}
The triplet  \( \hat{\gamma}_{i+1} \) with the highest probability is taken as the decoder prediction at step \( i + 1 \), and the process is repeated until termination criteria are reached.
Therefore, considering the semantic information extracted from questions and images, the problem $\mathcal{P}_1$ can be further converted as
\begin{subequations} \label{main_p3}
\begin{align}
    & \mathcal{P}_3: \max  \frac{1}{N_Q} \sum_{k=1}^{N_Q} \mathbbm{1} \{ f_{\text{ans}}(\Psi_k, G_{j_k} ;\theta_{\text{SE}},\theta_{\text{QP}},\theta_{\text{SG}} )  = A_k \}, \label{objective function_p3}\\
    & \text{s.t.}  ~~k \in [1, N_Q],~~ j_k \in [1, N_I], \label{p3_st_1}\\
    & ~~~~~~ W\leq W_0.\label{p3_st_2}
\end{align}
\end{subequations}

	
            
	       
			


\subsection{SG Semantic Ranker and GO-SG Algorithm}
We use a scene graph extraction model to extract the semantic information from the original image. The semantic information extraction process has two steps which are object identification and relationship capture. Specifically, a server first uses the model to detect, locate, and categorize all the objects in the image. Then, the model deduces the relationship between all the objects according to their geometry and logical correlation and outputs them in the form of triples.

We denote a directed graph by $G=<\mathcal{V},\mathcal{E},\mathcal{W}>$, where $\mathcal{V}$ represents a set of objectives (nodes), $\mathcal{E}\subseteq\mathcal{V}\times\mathcal{V}$ is a set of edges between the nodes indicating pairwise predicates (edges), and the function $\mathcal{W}: \mathcal{E}\leftarrow \bf{R}^+$ assigns each edge $e_{i,j}\in\mathcal{E}$ a positive weight, or zero if the edge does not exist.
Note that the weight can be obtained from the occurrence probability of the predicate.

The similarity between two SGs can be measured using graph edit distance (GED), a concept first formalized mathematically by \cite{sanfeliu1983distance}. This metric quantifies the minimum number of edit operations (e.g., node/edge insertion, deletion, or substitution) needed to transform one graph into another, and offers a way to quantify dissimilarity between graphs, similar to how string edit distance works for strings. Since the definition of GED depends on the properties of the graphs—such as whether the nodes and edges are labelled and whether the edges are directed—its formalization varies accordingly. In general, given a set of edit operations, the GED between two graphs $G_1$ and $G_2$ can be defined as
\begin{align}
    \text{GED}(G_1,G_2)=\min_{e_{\alpha,\alpha'}\in r(G_1,G_2)}\sum_{\alpha=1}^{k}c(e_{\alpha,\alpha'}),\label{ged}
\end{align}
where $\text{GED}(\cdot)$ is the graph edit operation function, $r(G_1,G_2)$ denotes the set of edit routes transforming $G_1$ to $G_2$, and $c(e_{\alpha,\alpha'})\geq 0$ is the cost of each graph edit operation $e_{\alpha,\alpha'}$.

Specifically, the set of elementary graph edit operators includes insertion, deletion, and substitution of nodes and edges.
\begin{align}
    \begin{split}
        c(e_{\alpha,\alpha'})= \left \{
	\begin{array}{ll}
    w^{\text{node}}_{\alpha},  &\alpha\neq \alpha' ,~~~ \alpha,\alpha'\in\mathcal{V};\\
	w^{\text{edge}}_{\alpha},  &\alpha \neq \alpha' ,~~~ \alpha,\alpha'\in\mathcal{E};\\
	0, & \text{otherwise}.
	\label{}
	\end{array}
	\right.
    \end{split}
\end{align}
Here, $w^{\text{node}}_{\alpha}$ and $w^{\text{edge}}_{\alpha}$ are the weights for node and edge edit, respectively.
 
Given that different tasks have different graphs, we can hence design the GO triplets transmission accordingly.
For each triplet $\phi_i \in G_{j_k}$, we define a ranking function as
\begin{align}
    f_{\text{score}}(\phi_i)= \text{GED}(\phi_i,G_{k}) \cdot \frac{n(\phi_i)}{\sum_{\phi_j\in G_{j_k}}n(\phi_j)}, \label{sg_score} 
\end{align}
where $n(\phi_i)$ is the occurrence frequency of the triplet $\phi_i$, and the second part $\frac{n(\phi_i)}{\sum_{\phi_j\in G_{j_k}}n(\phi_j)}$ is a padding weight to adjust the score considering the varying priority of different objects in the dataset. 
Therefore, we define the ranked semantic vector in SG as
\begin{equation}
\Psi_{j_k}^{\text{R,SG}}=\text{Rank}(f_{\text{score}}(G_{jk})), \label{ranked_sg}
\end{equation}
where Rank$(\cdot)$ represents a descending order sorting operation based on the calculated scores of the semantic triplets.
Then, we select the top $N_{\text{top}}$ entries in $G_{j_k}$, which are transmitted to the edge for answer reasoning, 
and the workflow of the proposed GO-SG algorithm is summarized in \textbf{Algorithm 2}.

\begin{algorithm}[t]
\caption{{GO-SG Algorithm}}
\label{algoritm_rank2}
 \begin{algorithmic}[1]
	\State Input:  $(Q_{k})_{k=1,...,N_Q}$, $(I_j)_{j=1,...,N_I}$.
    \State Input: Pre-trained question parser $\theta_{\text{QP}}$, semantic extractor $\theta_{\text{SE}}$, and SG generation $\theta_{\text{SG}}$.
	
    \For{$k = 1,...,N_Q$}
    	\State  Get question sequence $Q_{k}$.
\State  Generate the reasoning instruction steps $\Psi_k$ from question parser according to \eqref{after lstm}.
            
            \State  The end device receives the keywords $f_{\text{key}}(Q_{k})$ and captures the corresponding image  $ I_{j_k}$.
            \State  Extract the image semantic information $\Psi_{j_k}$ from the semantic extractor according to \eqref{after_cnn}.

            \State  Generate SG $G_{j_k}$ according to \eqref{after_cnn_sg}.

        \If{ $\phi_i \in G_{j_k}$}	        

                    \State Get the GO semantic ranking scores $f_{\text{score}}(\phi_i)$ for each triplet according to \eqref{sg_score}.
    \EndIf
          		
\EndFor			

\end{algorithmic}
\end{algorithm}

    \begin{figure*}[t!]
		\centering
		\subfigure[Accuracy for BBox methods in AWGN channel]{
	    \includegraphics[width=0.384\textwidth]{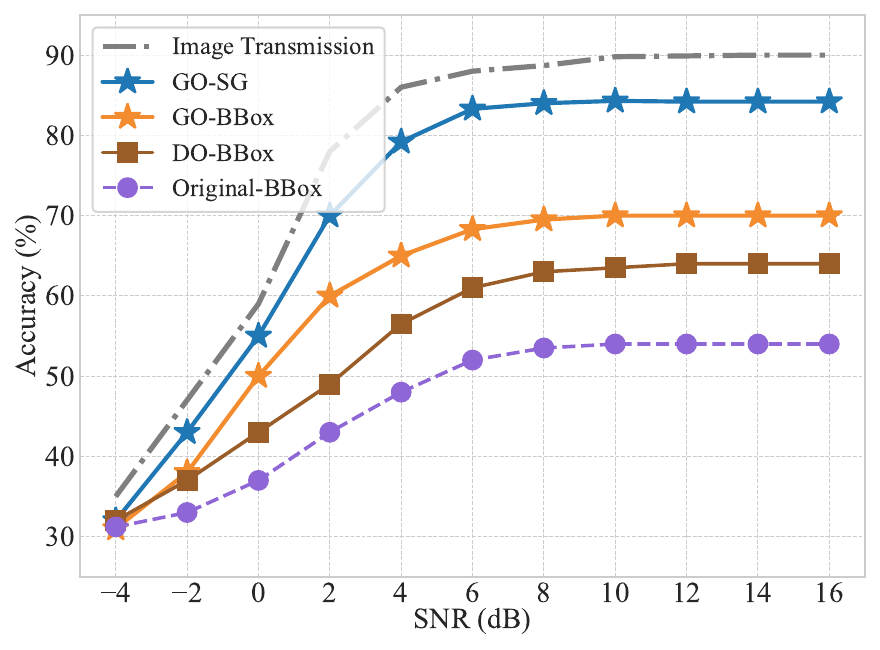}
		\label{range_snr_a}
		}
  \hspace{1.5cm}
  \subfigure[Accuracy for BBox methods in Rayleigh channel]{
	    \includegraphics[width=0.384\textwidth]{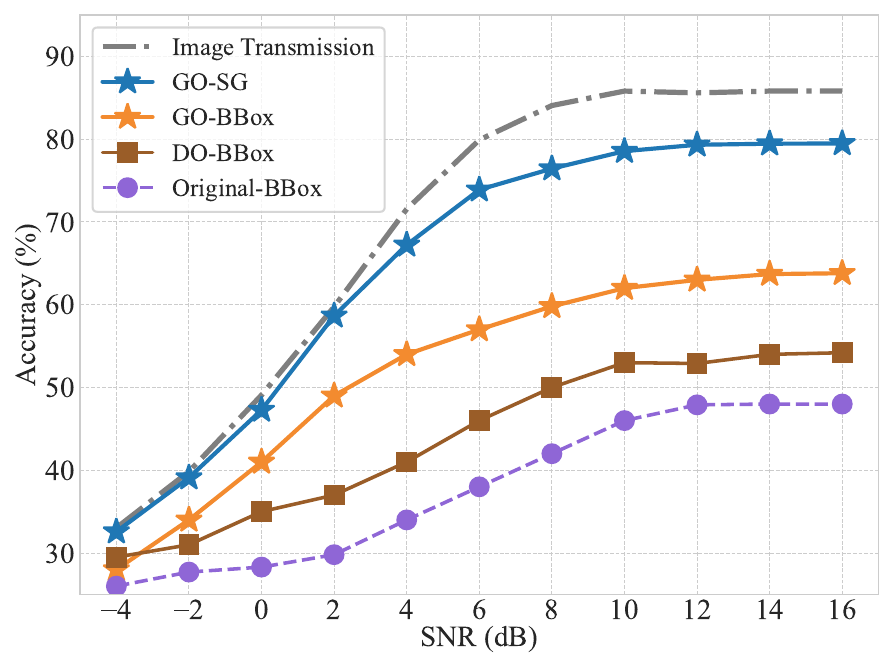}
		\label{range_snr_b}
		}
		\subfigure[Accuracy for SG methods in AWGN channel]{
		\includegraphics[width=0.384\textwidth]{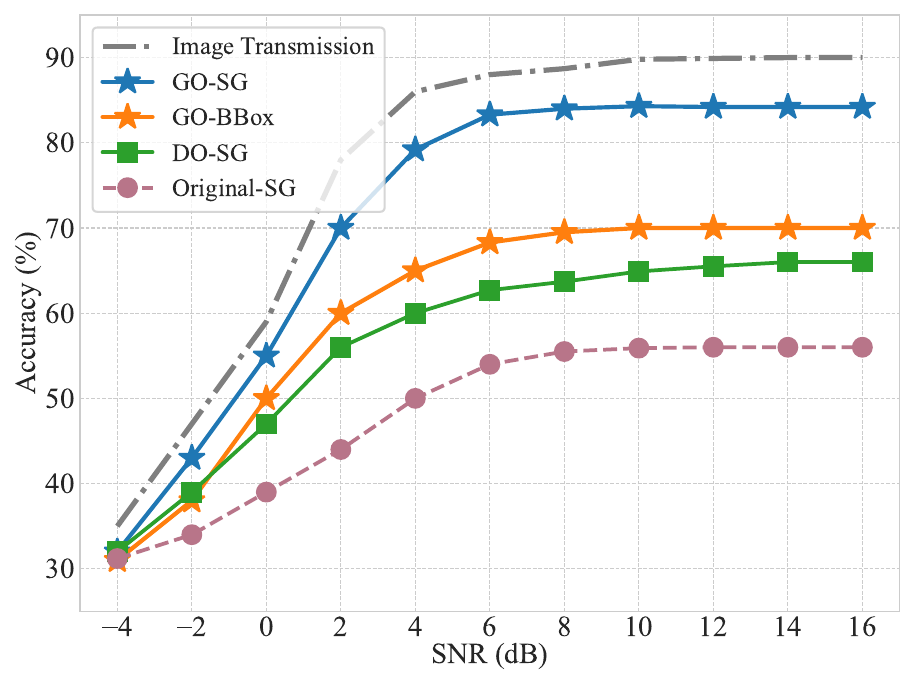}
		\label{range_snr_c}
		}
  \hspace{1.5cm}
  \subfigure[Accuracy for SG methods in Rayleigh channel]{
		\includegraphics[width=0.384\textwidth]{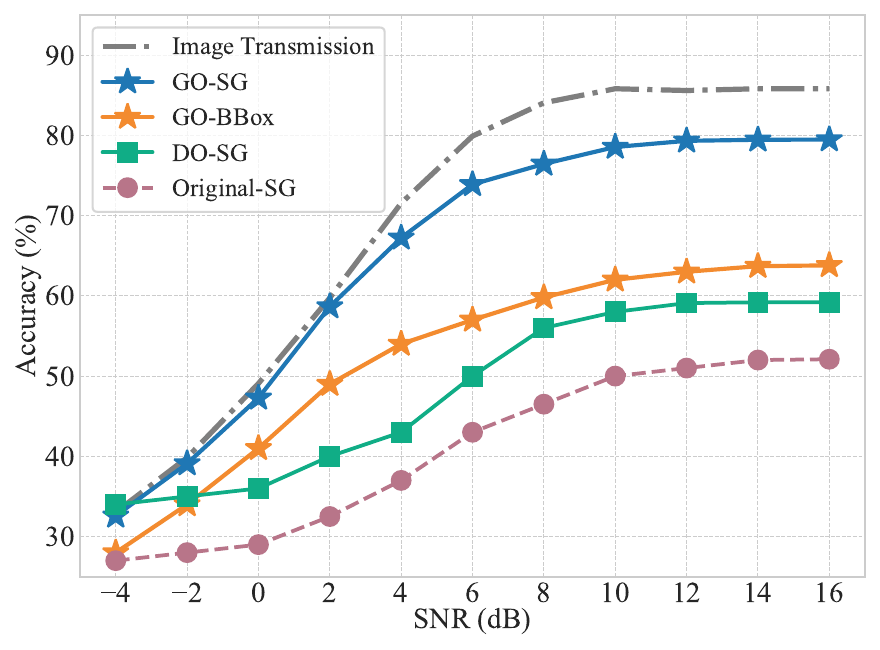}
		\label{range_snr_d}
		}
		\caption{Performance comparison in terms of accuracy in AWGN and Rayleigh channel.}
		\label{range_snr_all}
	\end{figure*}

\begin{figure*}[t!]
		\centering
		\subfigure[Performance comparison in AWGN channel]{
	    \includegraphics[width=0.36\textwidth]{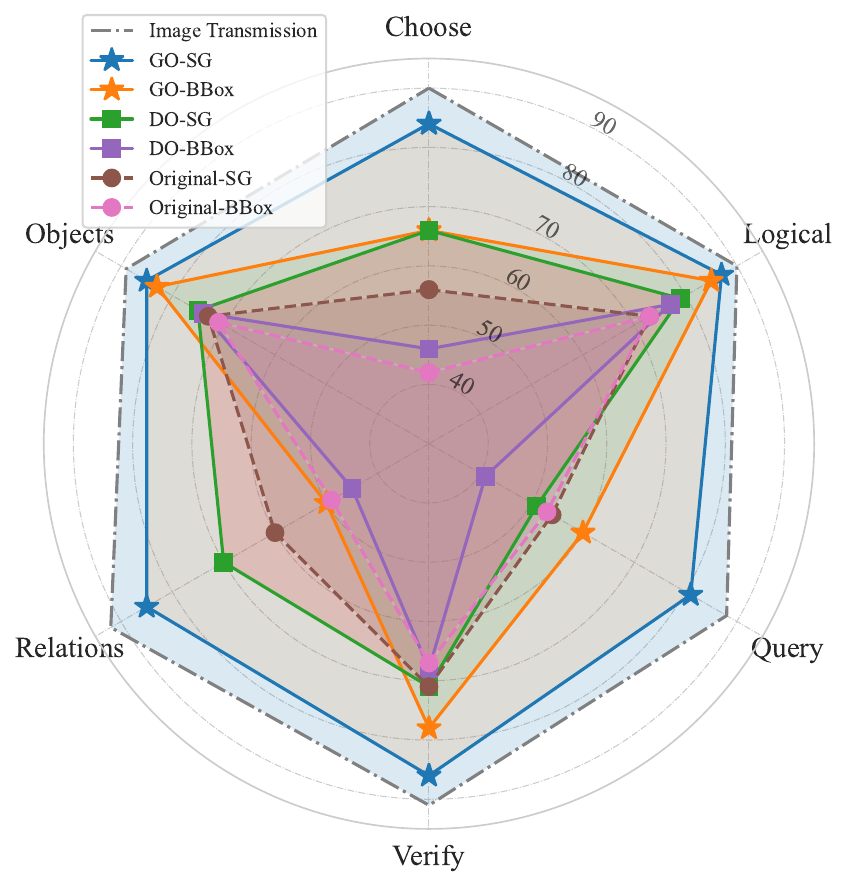}
		\label{performance_radar_a}
		}
  \subfigure[Performance comparison in Rayleigh channel]{
		\includegraphics[width=0.36\textwidth]{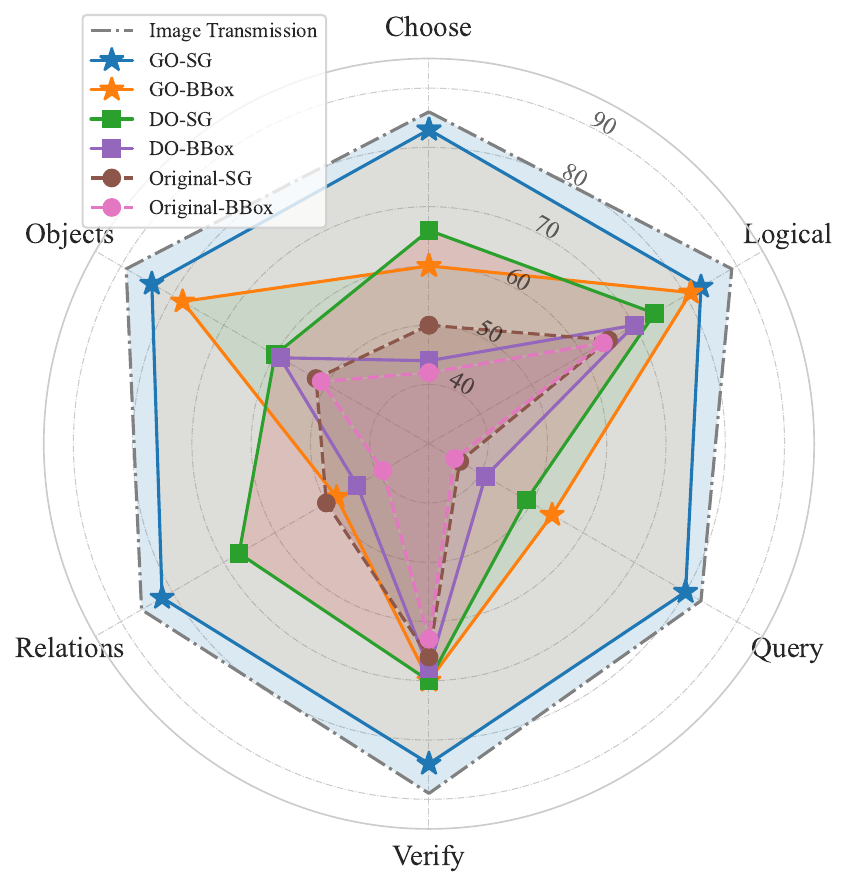}
		\label{performance_radar_b}
		}
		\caption{Performance comparison in terms of accuracy in AWGN and Rayleigh channel.}
		\label{performance_radar_all}
	\end{figure*}

\section{Simulation Results}
In this section, we compare our proposed GSC framework with traditional bit-oriented and data-oriented semantic methods, and we conduct different experiments to verify the superiority of the proposed GO-BBox and GO-SG algorithms.

\subsection{Experimental Settings}
We employ the GQV dataset \cite{hudson2019gqa} for our study, which focuses on real-world visual reasoning and compositional question answering, consisting of 22M questions over 110K images, which cover a wide range of reasoning skills and vary in length. The dataset has a vocabulary size of 3097 words, 1878 possible answers, 1704 classes and 311 relationships.
GQA is more challenging than other datasets (e.g., CLEVR \cite{johnson2017clevr} and VG \cite{krishna2017visualvq}) as all images are from the real world and each one is annotated with a dense scene graph and a large number of relations.

The Transformers has three encoder and decoder layers and twelve attention heads and it is trained based on cross-entropy loss and optimized using the Adam optimizer with learning rate $1\times10^{-3}$, batch size of $512$, and epoch of 100. 
The setting of the Mask R-CNN network with the ResNet-50 backbone network is the Adam optimizer with learning rate $5\times10^{-4}$, batch size of 256,  epoch of 100, and the input image pixel size and colour channels are $\text{H}\times \text{W}=480\times320$ and $\text{D}=3$.
We consider both AWGN and Rayleigh channel with bandwidth $W=100$ kHz, and the the varying value of SNR will be given in detail in the following results.
The experimental platform for this study employs Python 3.8.8 on the Ubuntu 22.04 system with two Nvidia Tesla A100, and PyTorch 2.1.1.

\begin{figure*}[t!]
		\centering
		\subfigure[Accuracy for BBox methods in AWGN channel]{
	    \includegraphics[width=0.36\textwidth]{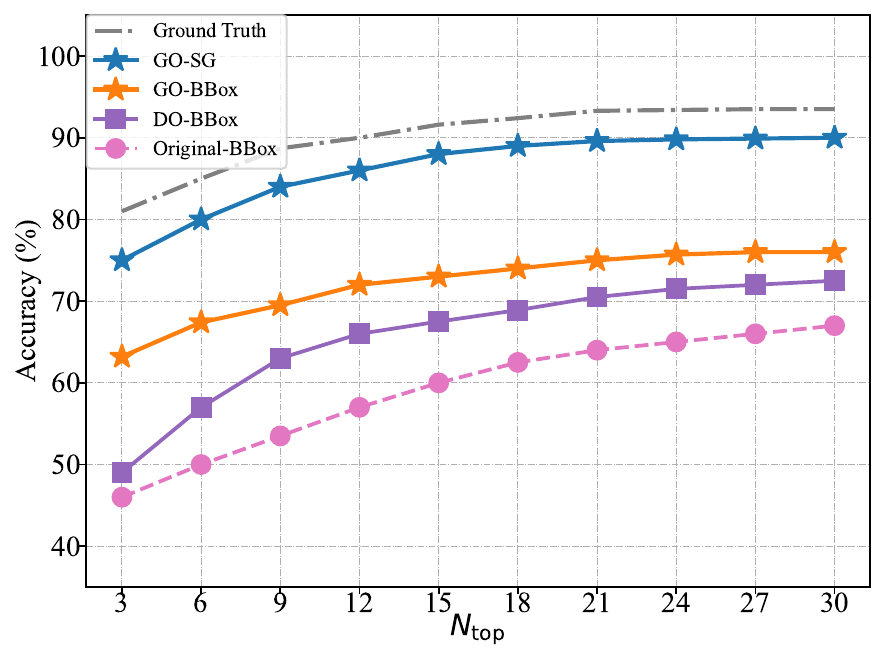}
		\label{range_n_a}
		}
  \subfigure[Accuracy for BBox methods in Rayleigh channel]{
	    \includegraphics[width=0.36\textwidth]{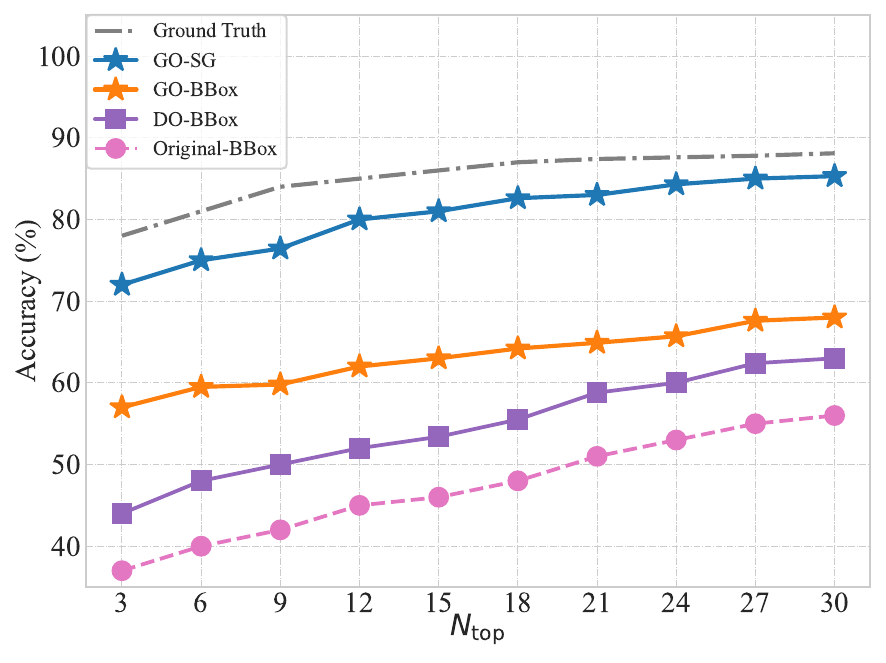}
		\label{range_n_b}
		}
\subfigure[Accuracy for SG methods in AWGN channel]{
		\includegraphics[width=0.36\textwidth]{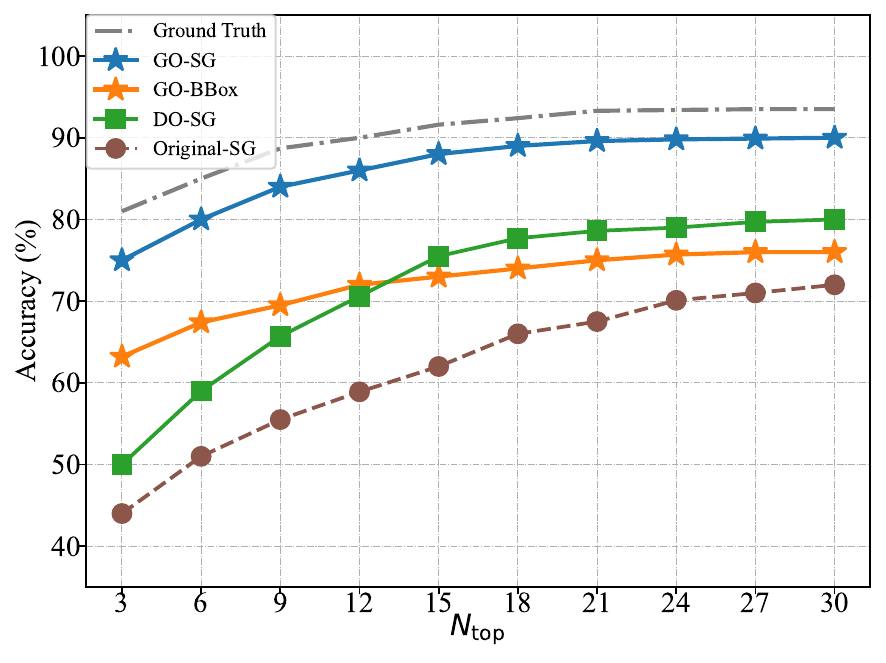}
		\label{range_n_c}
		}
  \subfigure[Accuracy for SG methods in Rayleigh channel]{
		\includegraphics[width=0.36\textwidth]{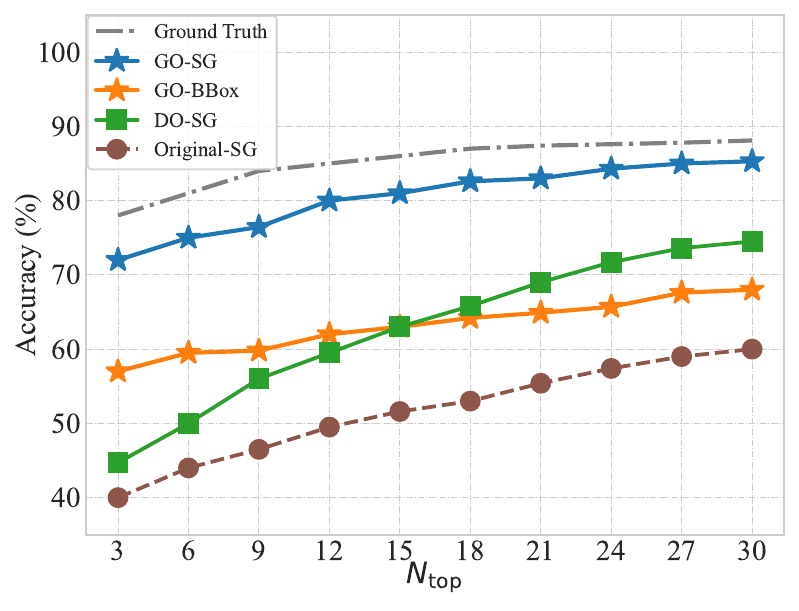}
		\label{range_n_d}
		}
		\caption{Performance comparison in terms of accuracy in AWGN and Rayleigh channel.}
		\label{range_n_all}
	\end{figure*}

\subsection{Methods for Comparison}
In order to better demonstrate the performance of our proposed algorithm GO-BBox and GO-SG, we introduce six comparative algorithms in the following.\textcolor{black}{
\begin{itemize}
    \item DO-SG. This algorithm is based on the data-oriented ranking method in \cite{zhang2023optimizationcmz}, ranking triplets based on their appearance frequency in the dataset without considering specific questions.
    \item     DO-BBox. Similar to DO-SG, this data-oriented algorithm ranks BBoxes solely based on their appearance frequency in the dataset.
    \item     Original-SG. This algorithm has no ranking process for SG, and the generated SG will be transmitted in the original ranking \cite{liang2021graghvqa}. 
    \item  Original-BBox. Similar to Original-SG, this algorithm has also no ranking process for BBox.
    \item  Ground Truth. 
    This algorithm assumes a noise-free transmission environment for SG transmission, serving as an ideal benchmark.
    \item     Image Transmission. 
    This bit-oriented approach transmits the entire image without semantic processing, assuming no time limitation for transmitting all data via the wireless channel to the edge server.    
\end{itemize}}
The works in \cite{zhang2023optimizationcmz} and \cite{liang2021graghvqa} focused solely on triplet execution.
To ensure a fair comparison with our proposed GSC methods, we extend these approaches by integrating additional components, including AWGN and Rayleigh channel simulations, channel encoders and decoders, a knowledge base, and LDPC error coding.
Furthermore, to align with the data format of SG methods, BBox methods are treated as triplets with empty relationships. 
The number of transmitted triplets or BBox elements, $N_{\text{top}}$, along with the SNR, varies across experiments to evaluate performance under different conditions.


  \begin{figure}[t]
  \centering
		\includegraphics[width=0.35\textwidth]{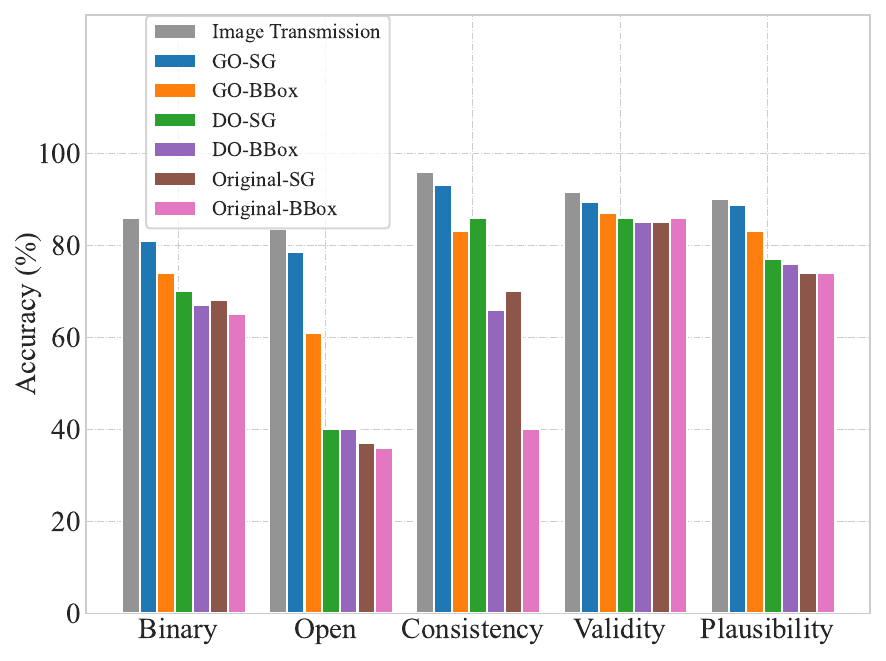}
		\caption{Performance evaluation on different dimensions}
        \label{performance_bar}
    \end{figure}

\subsection{Answering Accuracy for Various SNRs}
We plot the answering accuracy in Fig.~\ref{range_snr_all}, where both BBox-related and SG-related algorithms are evaluated in AWGN and Rayleigh channels. 
The Image Transmission algorithm serves as the performance upper bound since it has no time limitation to transmit data via the wireless channel.

%
As the SNR increases, the accuracy of all algorithms improves. Notably, our proposed GO-BBox and GO-SG algorithms show rapid improvement, particularly at lower SNRs (from -4 to 6 dB in AWGN and from -2 to 8 dB in Rayleigh), as they accurately extract the most relevant information needed to obtain the correct answer.
For both BBox and SG algorithms, the answer accuracy in the Rayleigh is worse than in the AWGN at equivalent SNR levels, due to the complex multipath effects in Rayleigh, whereas the simpler noise model in AWGN results in more stable transmission and higher accuracy.

Additionally, in Fig.~\ref{range_snr_a} and Fig.~\ref{range_snr_b}, the comparison of BBox-related algorithms shows that our proposed GO-BBox outperforms DO-BBox and Original-BBox in both AWGN and Rayleigh channels. 
This is because GO-BBox leverages goal-oriented information to select the most relevant BBox, leading to more accurate answers. 
In contrast, the DO-BBox algorithm relies solely on data-oriented information, allowing it to perform better than Original-BBox but worse than GO-BBox. 
Similarly, in Fig.~\ref{range_snr_c} and Fig.~\ref{range_snr_d}, which compare SG-related algorithms, GO-SG outperforms both DO-SG and Original-SG for same reasons related to the use of semantic information.  
It is noteworthy that, for GO, DO, and Original methods, SG algorithms generally outperform BBox algorithms, as SG captures a wider range of complex relationships, thereby enhancing accuracy.

   \begin{figure}[t]
		\centering
		\includegraphics[width=0.4\textwidth]{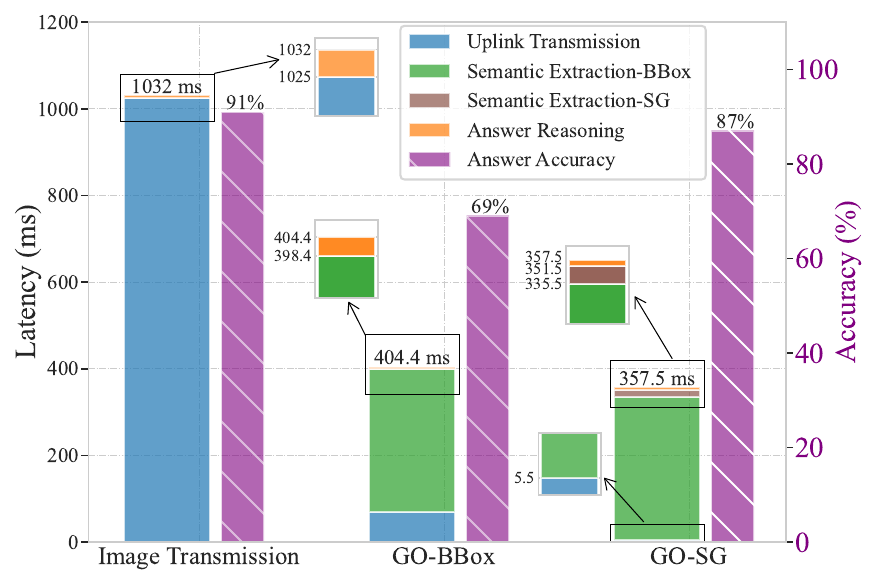}
		\caption{Latency comparison in full question-to-answer processing.}
        \label{latency_comparison}
    \end{figure}

 \begin{figure*}[!t]
		\centering
        \centering
		\includegraphics[width=0.9\textwidth]{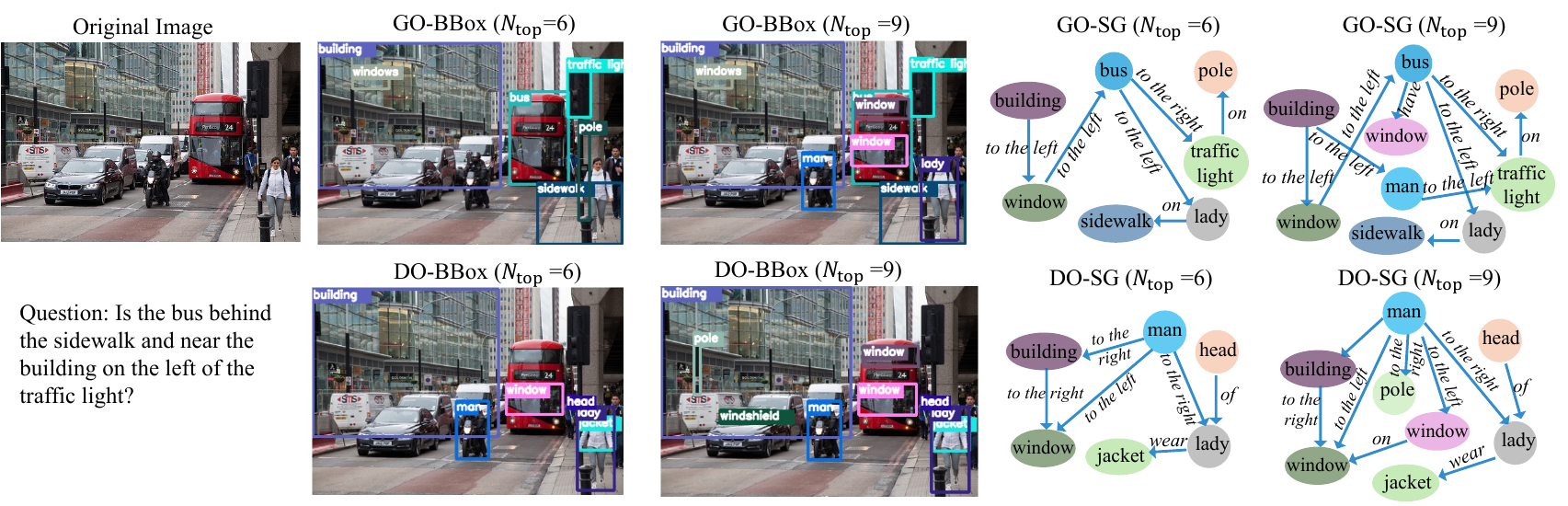}
		\caption{Example of Goal-Oriented Semantic Ranking for BBox and SG.}
        \label{fig_example}
    \end{figure*}

\subsection{Answering Accuracy for Various Question Types}
Fig.~\ref{performance_radar_all} presents the accuracy evaluation of various question types, comparing all BBox-based and SG-based algorithms in both AWGN and Rayleigh channels. The question types include ``Choose'', ``Logical'', ``Query'',``Verify'', ``Relations'', and ``Objects''. For this evaluation, the SNR is fixed at 8 dB and $N_{\text{top}}$ is set to 9.

The proposed GO-SG algorithm consistently outperforms other algorithms across all question types. 
Notably, for Query questions, GO-SG demonstrates a significant advantage over other methods. This is because Query questions are typically open-ended and require the algorithm for effective utilization of relationships and goal-oriented semantic triplets to reason the correct answer.
For Objects questions, BBox-based algorithms perform similarly to SG-based algorithms, as these questions mainly focus on identifying individual objects without involving complex relationships. 
In contrast, for Relations questions, SG algorithms outperform BBox algorithms because capturing and interpreting relationships between objects is crucial for achieving higher accuracy in these types of questions.


\subsection{Answering Accuracy for Various $N_{\text{top}}$}
Fig.~\ref{range_n_all} examines the impact of the number of selected triplets, $N_{\text{top}}$, on accuracy accuracy. The SNR is fixed at 8 dB for both AWGN and Rayleigh channels, and $N_{\text{top}}$ is varied from 3 to 30. 
The proposed GO-BBox and GO-SG methods outperform other comparable approaches. As $N_{\text{top}}$ increases, accuracy improves due to the transmission of more semantic triplets from the end device to the edge server. Additionally, in Fig.~\ref{range_n_c} and Fig.~\ref{range_n_d}, DO-SG performs worse than GO-BBox at small $N_{\text{top}}$ values (from 3 to 12 in AWN and from 3 to 15 in Rayleigh). 
This is because, when only a limited number of triplets are transmitted, the semantic information selected from the GO perspective is more critical for achieving accurate answers compared to that selected from the DO perspective, even in the absence of relationships in the BBox setting. 
However, after $N_{\text{top}} = 12$, DO-SG achieves higher accuracy as more triplets are transmitted, which allows additional relationships to be included, thereby increasing accuracy, even if some of them are not goal-oriented.



\subsection{Answer Accuracy for Various Evaluation Dimensions}
Fig.~\ref{performance_bar} examines the answering accuracy of our methods across five additional evaluation dimensions: ``Binary'', ``Open'', ``Consistency'', ``Validity'', and ``Plausibility'', as described in \cite{hudson2019gqa}. Binary and Open refer to yes/no questions and open-ended questions, respectively. The Consistency metric evaluates whether the answer is logically consistent with the questions. For example, ``A is to the left of B'' should be consistent with ``B is to the right of A''. Validity measures whether the provided answer is within the scope of the question, such as ensuring that a food/color/animal-related question receives a food/colour/animal-related answer. Plausibility assesses whether the answer is reasonable within the given context, such as ensuring that a response like ``buses wear hats'' is viewed as implausible.
In this analysis, we focus on the Rayleigh channel, with the SNR fixed at 8 dB and $N_{\text{top}}$ set to 9. It can be seen in the figure that our proposed GO-SG and GO-BBox algorithms outperform other SG and BBox methods across all five dimensions. 
Both algorithms perform well for open-ended questions due to the goal-oriented semantic ranking process, which preserves the most relevant information for these question types. For consistency-related questions, SG-based methods generally outperform their BBox counterparts, with DO-SG even surpassing GO-BBox, and Original-SG outperforming DO-BBox. This is because, in this scenario, capturing relationships plays a more critical role in ensuring consistency than goal-oriented ranking.



\subsection{Overall Latency Comparison}
\textcolor{black}{Fig.~\ref{latency_comparison} plots the overall latency comparison for Image Transmission, GO-BBox, and GO-SG, where lower latency and higher accuracy contribute to improved QoE for users. 
The latency for the entire process is calculated using equation \eqref{total_latency}, with the four specific latencies collectively contributing to the overall latency.
\begin{itemize}
\item  Uplink transmission refers to the communication latency involved in transmitting the bitstream data from the end device to the edge server over a Rayleigh channel (8 dB, 100 kHz). The latency is calculated using equation \eqref{t_tra} and is averaged across all image transmission computations.
\item Semantic Extraction-BBox represents the semantic extraction latency for BBox, calculated based on the Mask R-CNN-based BBox generation performed by the semantic extractor.
\item Semantic Extraction-SG corresponds to the semantic extraction latency for SG generation, derived based on the extracted BBox semantic information.
\item Answer Reasoning refers to the reasoning latency incurred by the answer reasoner at the edge server to derive the final answer.
\end{itemize}}
The complexity of BBox semantic extraction is 0.44 TFlops, and SG generation requires 0.02 TFlops, with the computations executed on an NVIDIA Jetson TX2 \cite{jetsontx2}, mounted on a DJI M210 drone \cite{djim210}, which has an AI performance of 1.33 TFlops. The answer reason latency is relatively small, accounting for only a minor portion of the overall process.

We can observe from the figure that Image Transmission suffers from significant bitstream transmission latency. For GO-BBox, the bitstream transmission latency is significantly reduced, although the computational latency for object detection increases considerably. Overall, the total latency decreases, but this reduction comes at the cost of lower answer accuracy. 
In the case of GO-SG, the latency associated with SG generation is added to the GO-BBox process, resulting in an increase in computation latency but a further reduction in bitstream transmission latency. Consequently, the total latency decreases marginally, while answer accuracy improves significantly, highlighting the semantic effectiveness and enhancing the QoE for the user.
Compared to Image Transmission, GO-SG significantly reduces the total latency (65\%) while incurring only a minimal loss in accuracy (4\%).
We provide a comprehensive comparison and practical guidelines for selecting the most suitable VQA communication mechanism. 


\subsection{Examples of Goal-Oriented Semantic Ranking}
In addition to the example shown in Fig.~\ref{fig_question} and Fig.~\ref{fig_image}, which highlight the differences between BBox and SG by demonstrating SG's ability to capture more relationships, we present another example from a goal-oriented perspective to compare GO-SG, GO-BBox, DO-SG, and DO-BBox for $N_{\text{top}} = 6$ and 9, respectively, in Fig.\ref{fig_example}.
This example includes an original image and a corresponding question: ``Is the bus behind the sidewalk and near the building to the left of the traffic light?" We observe that both GO-BBox and GO-SG can semanticly extract the BBox and SG relevant to the question. 
In contrast, DO-BBox and DO-SG do not achieve the same precision because they focus on data-oriented semantic ranking and neglect the goal of answering the specific question. 
This disparity is reduced when $N_{\text{top}} = 9$, as more triplets are available for transmission. However, when the number of transmittable triplets is limited ($N_{\text{top}} = 6$), the performance gap between GO and DO becomes much more pronounced.

\section{Conclusions}
In this paper, we proposed a novel goal-oriented semantic communication (GSC) framework designed to enhance the performance and efficiency of Visual Question Answering (VQA) systems in edge-enabled wireless scenarios. 
Our proposed framework effectively addresses the challenges of limited communication resources and noisy channels by prioritizing the transmission of semantically significant information most relevant to answering VQA question.
To achieve this, we developed a goal-oriented bounding box (GO-BBox) ranking approach and extended it with a scene graph (GO-SG) ranking method to handle complex relationship-based questions. These methods leverage advanced semantic extraction and prioritization strategies to optimize the transmission of relevant information under resource-constrained conditions.
Extensive experiments under AWGN and Rayleigh channels demonstrated that the proposed GSC framework significantly outperforms traditional bit-oriented and other existing semantic communication approaches. 
Our results demonstrated substantial improvements in answering accuracy, up to 49\% in AWGN channels and 59\% in Rayleigh channels, while reducing total execution latency by up to 65\%. 
These results validated the effectiveness of our proposed GSC framework in enhancing VQA answering accuracy under diverse channel conditions and resource constraints, while offering practical guidelines for selecting the most suitable communication mechanism to ensure adaptability to varying requirements.

\bibliography{JSAC}

\begin{thebibliography}{10}
\providecommand{\url}[1]{#1}
\csname url@samestyle\endcsname
\providecommand{\newblock}{\relax}
\providecommand{\bibinfo}[2]{#2}
\providecommand{\BIBentrySTDinterwordspacing}{\spaceskip=0pt\relax}
\providecommand{\BIBentryALTinterwordstretchfactor}{4}
\providecommand{\BIBentryALTinterwordspacing}{\spaceskip=\fontdimen2\font plus
\BIBentryALTinterwordstretchfactor\fontdimen3\font minus \fontdimen4\font\relax}
\providecommand{\BIBforeignlanguage}[2]{{%
\expandafter\ifx\csname l@#1\endcsname\relax
\typeout{** WARNING: IEEEtran.bst: No hyphenation pattern has been}%
\typeout{** loaded for the language `#1'. Using the pattern for}%
\typeout{** the default language instead.}%
\else
\language=\csname l@#1\endcsname
\fi
#2}}
\providecommand{\BIBdecl}{\relax}
\BIBdecl

\bibitem{antol2015vqa}
S.~Antol, A.~Agrawal, J.~Lu, M.~Mitchell, D.~Batra, C.~L. Zitnick, and D.~Parikh, ``{VQA}: Visual question answering,'' in \emph{Proceedings of the IEEE Int. Conf. Comp. Vis. (ICCV)}, Santiago, Chile, December 2015, pp. 2425--2433.

\bibitem{lu2016hierarchical}
J.~Lu, J.~Yang, D.~Batra, and D.~Parikh, ``Hierarchical question-image co-attention for visual question answering,'' \emph{Adv. Neural Inf. Process. Syst. (NeurIPS)}, vol.~29, 2016.

\bibitem{shi2016edge}
W.~{Shi}, J.~{Cao}, Q.~{Zhang}, Y.~{Li}, and L.~{Xu}, ``Edge computing: Vision and challenges,'' \emph{IEEE Internet Things J.}, vol.~3, no.~5, pp. 637--646, Oct. 2016.

\bibitem{9760192lsg}
S.~Liu, P.~Cheng, Z.~Chen, W.~Xiang, B.~Vucetic, and Y.~Li, ``Contextual user-centric task offloading for mobile edge computing in ultra-dense network,'' \emph{IEEE Trans. Mobile Comput.}, vol.~22, no.~9, pp. 5092--5108, Sep. 2023.

\bibitem{luo2022semantic}
X.~Luo, H.-H. Chen, and Q.~Guo, ``Semantic communications: Overview, open issues, and future research directions,'' \emph{IEEE Trans. Wireless Commun.}, vol.~29, no.~1, pp. 210--219, Feb. 2022.

\bibitem{10644029zhouhui}
H.~Zhou, Y.~Deng, X.~Liu, N.~Pappas, and A.~Nallanathan, ``Goal-oriented semantic communications for 6g networks,'' \emph{IEEE Internet Things Mag.}, vol.~7, no.~5, pp. 104--110, Sep. 2024.

\bibitem{yan2022resource}
L.~Yan, Z.~Qin, R.~Zhang, Y.~Li, and G.~Y. Li, ``Resource allocation for text semantic communications,'' \emph{IEEE Wireless Commun. Lett.}, vol.~11, no.~7, pp. 1394--1398, Jul. 2022.

\bibitem{weng2021semantic}
Z.~Weng, Z.~Qin, and G.~Y. Li, ``Semantic communications for speech signals,'' in \emph{Proc. IEEE Int. Conf. Commun. (ICC)}.\hskip 1em plus 0.5em minus 0.4em\relax IEEE, 2021, pp. 1--6.

\bibitem{9928407}
A.~Li, X.~Liu, G.~Wang, and P.~Zhang, ``Domain knowledge driven semantic communication for image transmission over wireless channels,'' \emph{IEEE Wireless Commun. Lett.}, vol.~12, no.~1, pp. 55--59, Jan. 2023.

\bibitem{jiang2022wireless}
P.~Jiang, C.-K. Wen, S.~Jin, and G.~Y. Li, ``Wireless semantic communications for video conferencing,'' \emph{IEEE J. Sel. Areas Commun.}, vol.~41, no.~1, pp. 230--244, Jan. 2022.

\bibitem{bourtsoulatze2019deepjscc}
E.~Bourtsoulatze, D.~B. Kurka, and D.~G{\"u}nd{\"u}z, ``Deep joint source-channel coding for wireless image transmission,'' \emph{IEEE Trans. Cogn. Commun. Netw.}, vol.~5, no.~3, pp. 567--579, Sep. 2019.

\bibitem{xu2021wireless}
J.~Xu, B.~Ai, W.~Chen, A.~Yang, P.~Sun, and M.~Rodrigues, ``Wireless image transmission using deep source channel coding with attention modules,'' \emph{IEEE Trans. Circuits Syst. Video Technol.}, vol.~32, no.~4, pp. 2315--2328, Apr. 2021.

\bibitem{9878262}
H.~Wu, Y.~Shao, K.~Mikolajczyk, and D.~Gündüz, ``Channel-adaptive wireless image transmission with {OFDM},'' \emph{IEEE Wireless Commun. Lett.}, vol.~11, no.~11, pp. 2400--2404, Nov. 2022.

\bibitem{diwan2023objectyolo}
T.~Diwan, G.~Anirudh, and J.~V. Tembhurne, ``Object detection using yolo: Challenges, architectural successors, datasets and applications,'' \emph{multimedia Tools and Applications}, vol.~82, no.~6, pp. 9243--9275, 2023.

\bibitem{Girshick_2015_ICCVfasterrcnn}
R.~Girshick, ``{Fast R-CNN},'' in \emph{Proc. IEEE Int. Conf. Comput. Vis. (ICCV)}, Santiago, Chile, December 2015, pp. 1440--1448.

\bibitem{he2017mask}
K.~He, G.~Gkioxari, P.~Doll{\'a}r, and R.~Girshick, ``Mask {R-CNN},'' in \emph{Proc. IEEE Int. Conf. Comput. Vis. (ICCV)}, Venice, Italy, Oct. 2017, pp. 2961--2969.

\bibitem{yi2018neural}
K.~Yi, J.~Wu, C.~Gan, A.~Torralba, P.~Kohli, and J.~Tenenbaum, ``Neural-symbolic {VQA}: Disentangling reasoning from vision and language understanding,'' \emph{Adv. Neural Inf. Process. Syst. (NeurIPS)}, vol.~31, 2018.

\bibitem{vatashsky2020vqa}
B.-Z. Vatashsky and S.~Ullman, ``{VQA} with no questions-answers training,'' in \emph{Proc. IEEE/CVF Conf. Comput. Vis. Pattern Recog. (CVPR)}, 2020, pp. 10\,376--10\,386.

\bibitem{teney2017graph}
D.~Teney, L.~Liu, and A.~van Den~Hengel, ``Graph-structured representations for visual question answering,'' in \emph{Proc. IEEE Conf. Comput. Vis. Pattern Recog. (CVPR)}, 2017, pp. 1--9.

\bibitem{li2019relation}
L.~Li, Z.~Gan, Y.~Cheng, and J.~Liu, ``Relation-aware graph attention network for visual question answering,'' in \emph{Proc. IEEE/CVF Conf. Comput. Vis. Pattern Recog. (CVPR)}, 2019, pp. 10\,313--10\,322.

\bibitem{zellers2018neuralmotif}
R.~Zellers, M.~Yatskar, S.~Thomson, and Y.~Choi, ``Neural motifs: Scene graph parsing with global context,'' in \emph{Proc. IEEE Conf. Comput. Vis. Pattern Recog. (CVPR)}, 2018, pp. 5831--5840.

\bibitem{yang2018graphrcnn}
J.~Yang, J.~Lu, S.~Lee, D.~Batra, and D.~Parikh, ``Graph r-cnn for scene graph generation,'' in \emph{Proc. Eur. Conf. Comput. Vis. (ECCV)}, 2018, pp. 670--685.

\bibitem{tang2020unbiased}
K.~Tang, Y.~Niu, J.~Huang, J.~Shi, and H.~Zhang, ``Unbiased scene graph generation from biased training,'' in \emph{Proc. IEEE/CVF Conf. Comput. Vis. Pattern Recog. (CVPR)}, 2020, pp. 3716--3725.

\bibitem{khandelwal2022iterative}
S.~Khandelwal and L.~Sigal, ``Iterative scene graph generation,'' \emph{Adv. Neural Inf. Process. Syst. (NeurIPS)}, vol.~35, pp. 24\,295--24\,308, 2022.

\bibitem{xie2021task}
H.~Xie, Z.~Qin, and G.~Y. Li, ``Task-oriented multi-user semantic communications for {VQA},'' \emph{IEEE Wireless Commun. Lett.}, vol.~11, no.~3, pp. 553--557, Dec. 2021.

\bibitem{10255282}
Z.~Q. Liew, M.~Xu, W.~Y.~B. Lim, Z.~Xiong, D.~Niyato, and D.~I. Kim, ``Mechanism design for semantic communication in uav-assisted metaverse: A combinatorial auction approach,'' \emph{IEEE Trans. Veh. Technol.}, vol.~73, no.~2, pp. 2236--2251, Feb. 2024.

\bibitem{zhang2023optimizationcmz}
W.~Zhang, Y.~Wang, M.~Chen, T.~Luo, and D.~Niyato, ``Optimization of image transmission in a cooperative semantic communication networks,'' \emph{IEEE Trans. Wireless Commun.}, vol.~23, no.~2, Feb. 2023.

\bibitem{10618994wenchao}
W.~Wu, Y.~Yang, Y.~Deng, and A.~Hamid~Aghvami, ``Goal-oriented semantic communications for robotic waypoint transmission: The value and age of information approach,'' \emph{IEEE Trans. Wireless Commun.}, pp. 1--1, 2024.

\bibitem{10577270zhe}
Z.~Wang, Y.~Deng, and A.~Hamid~Aghvami, ``Goal-oriented semantic communications for avatar-centric augmented reality,'' \emph{IEEE Trans. Commun.}, pp. 1--1, 2024.

\bibitem{liang2021graghvqa}
W.~Liang, Y.~Jiang, and Z.~Liu, ``{GraghVQA}: Language-guided graph neural networks for graph-based visual question answering,'' \emph{arXiv preprint arXiv:2104.10283}, 2021.

\bibitem{chen2005reduced}
J.~Chen, A.~Dholakia, E.~Eleftheriou, M.~P. Fossorier, and X.-Y. Hu, ``Reduced-complexity decoding of {LDPC} codes,'' \emph{IEEE Trans. Commun.}, vol.~53, no.~8, pp. 1288--1299, Aug. 2005.

\bibitem{pennington2014glove}
J.~Pennington, R.~Socher, and C.~D. Manning, ``Glove: Global vectors for word representation,'' in \emph{Proc. 2014 Conf. Empir. Methods Nat. Lang. Process. (EMNLP)}, 2014, pp. 1532--1543.

\bibitem{vaswani2017attention}
A.~Vaswani, ``Attention is all you need,'' \emph{Adv. Neural Inf. Process. Syst. (NeurIPS)}, pp. 6000--6010, 2017.

\bibitem{sanfeliu1983distance}
A.~Sanfeliu and K.-S. Fu, ``A distance measure between attributed relational graphs for pattern recognition,'' \emph{IEEE Trans. Syst. Man Cybern.}, no.~3, pp. 353--362, 1983.

\bibitem{hudson2019gqa}
D.~A. Hudson and C.~D. Manning, ``{GQA}: A new dataset for real-world visual reasoning and compositional question answering,'' in \emph{Proc. IEEE/CVF Conf. Comput. Vis. Pattern Recog. (CVPR)}, Long Beach, USA, 2019, pp. 6700--6709.

\bibitem{johnson2017clevr}
J.~Johnson, B.~Hariharan, L.~Van Der~Maaten, L.~Fei-Fei, C.~Lawrence~Zitnick, and R.~Girshick, ``{CLEVR}: A diagnostic dataset for compositional language and elementary visual reasoning,'' in \emph{Proc. IEEE Conf. Comput. Vis. Pattern Recog. (CVPR)}, Honolulu, USA, 2017, pp. 2901--2910.

\bibitem{krishna2017visualvq}
R.~Krishna, Y.~Zhu, O.~Groth, J.~Johnson, K.~Hata, J.~Kravitz, S.~Chen, Y.~Kalantidis, L.-J. Li, D.~A. Shamma \emph{et~al.}, ``Visual genome: Connecting language and vision using crowdsourced dense image annotations,'' \emph{Int. J. Comput. Vis.}, vol. 123, pp. 32--73, 2017.

\bibitem{jetsontx2}
NVIDIA, ``{NVIDIA Jetson TX2 Next-level performance for mass-market AI products},'' 2024, [Online]. Available \url{https://www.nvidia.com/en-gb/autonomous-machines/embedded-systems/jetson-tx2/}.

\bibitem{djim210}
DJI, ``M210 onboard computer checklis,'' 2024, [Online]. Available \url{https://developer.dji.com/onboard-sdk/documentation/M210-Docs/oes-checklist.html}.

\end{thebibliography}
\bibliographystyle{IEEEtran}

\end{document}